\documentclass{article}

    \PassOptionsToPackage{numbers, compress}{natbib}

\usepackage[preprint]{neurips_2026}

\usepackage[utf8]{inputenc} 
\usepackage[T1]{fontenc}    
\usepackage{hyperref}       
\usepackage{url}            
\usepackage{booktabs}       
\usepackage{amsfonts}       
\usepackage{nicefrac}       
\usepackage{microtype}      
\usepackage{xcolor}         
\usepackage{subcaption}

\usepackage{algorithm}
\usepackage{algpseudocode}
\usepackage{amsmath}
\usepackage{amssymb}
\usepackage{multirow}
\usepackage{enumitem}
\usepackage[table]{xcolor}
\definecolor{oursrow}{RGB}{235,250,235}
\newcommand{\std}[1]{\raisebox{0.1ex}{\scriptsize$\pm$#1}}

\usepackage[textsize=tiny]{todonotes}
\newcommand{\system}{\texttt{MedMix}}

\title{\system{}: Specialization-Consistent Federated \\ Sparse MoEs under Modality Heterogeneity}

\author{
Adiba Orzikulova$^{1}$\; Dong Min Kim$^{1}$\;
Jaehong Yoon$^2$\; Sung-Ju Lee$^1$\; \\
 $^1$KAIST, $^2$NTU Singapore \\
    \texttt{\{adiorz,dongmin.kim,profsj\}@kaist.ac.kr}\\  
    \texttt{jaehong.yoon@ntu.edu.sg}
}
\begin{document}

\maketitle

\begin{abstract}
    Federated multimodal medical AI faces modality heterogeneity at both the client and sample levels: clients may systematically lack access to specific modality types, while individual records within the same client may contain different partial modality subsets.
    Sparse Mixture-of-Experts~(MoE) architectures are a promising remedy for modality-adaptive computation, but their use in federated learning is fragile under cross-client modality heterogeneity, where locally learned routing policies can diverge across clients and drive experts toward incompatible specializations. Different clients may assign the same observed modality configuration to different experts, or train similarly indexed experts on different missing-modality configurations, causing standard aggregation to misalign or overwrite the expert specialization that sparse MoEs are intended to learn. To address this challenge, we propose \system{}, 
    a semantic-alignment framework for federated multimodal sparse MoEs that coordinates cross-client routing and expert specialization using modality context.
    At the client side, \system{} uses \emph{modality-context-aware routing} to guide expert selection using each token's modality identity, position, and incompleteness context. Across clients, it uses \emph{consensus-guided routing alignment} to construct server-side consensus anchors for shared modality patterns and align local routing distributions across clients.
    Complementing these routing mechanisms, \emph{client-adaptive expert aggregation} leverages client-specific modality-pattern prototypes to match and aggregate functionally similar experts across clients. Experiments on real-world multimodal medical datasets show that \system{} achieves the best average F1 across diverse modality heterogeneity and modality incompleteness settings, with especially clear gains under severe heterogeneity.
\end{abstract}


\section{Introduction}\label{s:introduction}
In medical AI, multimodal data such as MRI images, lab tests, and clinical text reports enable more accurate and robust inference by capturing complementary evidence about a patient's state~\citep{goldberger2000physiobank}. This capability is especially important in real-world clinical settings, ranging from Alzheimer’s disease diagnosis~\citep{weiner2010alzheimer, weiner2017alzheimer} to ICU outcome prediction using heterogeneous electronic health records~\citep{johnson2023mimic, johnson2023mimic_note, johnson2024mimiciv_dataset}. Yet deploying multimodal models across institutions remains constrained by two key challenges.

First, medical data are \emph{privacy-sensitive} and therefore remain siloed within individual institutions. Regulations such as HIPAA~(Health Insurance Portability and Accountability Act) and GDPR~(General Data Protection Regulation) make centralizing raw patient records legally and operationally impractical, motivating federated learning (FL)~\citep{mcmahan2017communication, kairouz2021advances} as the default collaboration paradigm~\citep{nguyenlearning, orzikulova2024federated, orzikulova2025fedduet}. 
Second, clinical records exhibit pervasive modality incompleteness~\citep{johnson2023mimic_note, johnson2023mimic, weiner2010alzheimer}.
At the institution level, hospitals may differ in which modalities are routinely available because of differences in equipment, workflows, and clinical practice. At the patient level, individual records may contain only a subset of modalities because tests are not ordered, scans are not acquired, or notes are unavailable. These factors induce \emph{dual-level modality heterogeneity}~\citep{orzikulova2025fedduet}, consisting of cross-client differences in modality availability and within-client variation in patient-level incompleteness.

Sparse Mixture-of-Experts~(MoE) architectures~\citep{shazeer2017outrageously,lepikhin2020gshard,fedus2022switch} are appealing for multimodal learning under modality incompleteness because they can allocate different inputs to specialized experts.
Recent multimodal MoE methods have shown strong robustness to arbitrary or incomplete modality combinations in centralized settings~\citep{yun2024flex,han2024fusemoe}. However, their interaction with FL aggregation is fragile under dual-level modality heterogeneity. When clients observe different modality combinations with different frequencies, local training induces different routing signals across clients. This can cause routers to develop client-specific expert preferences, even for modality configurations shared across clients. Moreover, these divergent routing patterns shape how experts are trained locally: experts with the same index may receive inputs from different modality contexts on different clients. Consequently, same-index experts can acquire different functional roles, weakening the semantic compatibility of expert specialization required for direct aggregation. This weakens the benefit of federated aggregation, leading to degraded performance under modality heterogeneity (see Figure~\ref{fig:motivation_diagnostics}).

We propose \system{}, a semantic-alignment framework for federated multimodal sparse MoEs that coordinates cross-client routing and expert specialization under modality heterogeneity. 
\system{} consists of three complementary components:
(1) \emph{modality-context-aware routing} guides expert selection using each token's modality identity, position, and incompleteness context;
(2) \emph{consensus-guided routing alignment} constructs reliable consensus routing anchors for shared modality patterns and aligns local batch-level routing distributions to these anchors; and
(3) \emph{client-adaptive expert aggregation} uses client-specific modality-pattern prototypes to match and aggregate functionally similar experts while preserving client-specific specialization.
By coupling these mechanisms, \system{} promotes semantically consistent routing and expert specialization across clients while adapting aggregation to heterogeneous modality distributions.

We evaluate \system{} on two real-world multimodal medical benchmarks: ADNI~\citep{weiner2010alzheimer}, integrating MRI, genomic, clinical, and biospecimen modalities for Alzheimer’s disease assessment, and MIMIC-IV~\citep{johnson2023mimic,johnson2023mimic_note}, integrating laboratory measurements, clinical notes, and diagnosis codes for ICU outcome prediction.
Both datasets exhibit natural modality incompleteness. To systematically study federated heterogeneity, we construct splits with varying degrees of inter-client modality availability shift and intra-client sample-level modality incompleteness, including an extreme modality-exclusive setting.
Across these scenarios, \system{} achieves the best average F1 score, with larger gains under stronger modality heterogeneity. These results highlight the importance of cross-client specialization consistency through stabilized routing and heterogeneity-aware expert aggregation in federated multimodal MoEs. Our contributions are summarized as follows:
\begin{itemize}[leftmargin=2em, itemsep=0.2em, topsep=0.2em]
    \item We identify a key fragility of federated multimodal MoEs under dual-level modality heterogeneity, where client-specific modality distributions induce inconsistent routing behavior and expert specialization across clients, weakening the semantic compatibility of expert specialization required for effective federated aggregation.
    \item We propose \system{}, a federated multimodal sparse MoE framework that promotes semantic consistency in routing and expert specialization across clients while adapting aggregation to modality heterogeneity.
    \item We evaluate \system{} on ADNI and MIMIC-IV datasets across diverse modality heterogeneity scenarios, demonstrating consistent improvements over strong federated multimodal and MoE baselines, particularly under severe modality heterogeneity.
\end{itemize}

\vspace{-2mm}
\section{Background}\label{sec:background}
\begin{figure*}[t]
  \centering
  \includegraphics[width=\textwidth]{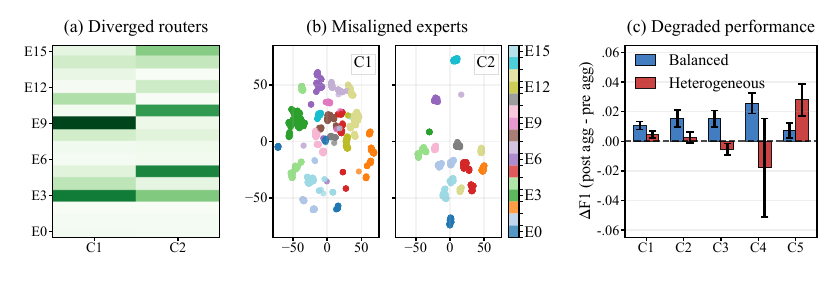}
  \caption{\textbf{Fragility of federated MoE specialization under modality heterogeneity.}
  (a) \textbf{Routing divergence:} normalized top-$k$ expert activation frequencies. Clients show divergent routing preferences under the same observed modality configuration. 
  (b) \textbf{Expert misalignment:} joint t-SNE of expert-layer outputs from two clients, with points colored by the top-1 routed expert. Rearranged color and cluster structure indicates that same-index experts acquire different client-specific semantic roles.
  (c) \textbf{Aggregation impact:} validation F1 change after aggregation relative to the pre-aggregation local model, $\Delta\mathrm{F1}=\mathrm{F1}_{\mathrm{post}}-\mathrm{F1}_{\mathrm{pre}}$,
  averaged over communication rounds and 5 seeds. 
  Aggregation is beneficial in the balanced split but becomes less reliable under heterogeneous modality distributions, reflecting reduced cross-client specialization consistency.}
  \label{fig:motivation_diagnostics}
\end{figure*}

\paragraph{Multimodal Federated Learning.}
We study supervised multimodal federated learning (MFL) under modality incompleteness.
There are $K$ clients and $M$ possible modalities.
Client $k$ owns
$\mathcal{D}_k=\{(\mathbf{x}_{i,k},y_{i,k},\mathbf{r}_{i,k})\}_{i=1}^{n_k}$,
where $\mathbf{x}_{i,k}=\{x^m_{i,k}\}_{m=1}^{M}$ is the multimodal input,
$y_{i,k}$ is the label, and
$\mathbf{r}_{i,k}\in\{0,1\}^{M}$ is the modality-availability mask indicating which modalities are observed. $\Omega(\mathbf{r})=\{m:r^m=1\}$ denotes observed modality set. The goal is to learn a global model $f_\theta$: 
\begin{equation}
    \min_{\theta}
    \sum_{k=1}^{K} \frac{n_k}{n}
    \mathbb{E}_{(\mathbf{x},y,\mathbf{r})\sim\mathcal{D}_k}
    \left[
        \ell\left(f_\theta(\mathbf{x},\mathbf{r}),y\right)
    \right],
    \qquad
    n=\sum_{k=1}^{K} n_k .
    \label{eq:mfl_objective}
\end{equation}
Here, $f_\theta(\mathbf{x},\mathbf{r})$ denotes prediction using only the modalities in $\Omega(\mathbf{r})$.

\paragraph{Sparse Mixture of Experts.}
Sparse Mixture-of-Experts~(MoE) models increase model capacity while activating only a sparse subset of parameters for each input~\citep{shazeer2017outrageously}.
Recent multimodal MoE methods further show that sparse expert routing is effective under modality incompleteness and arbitrary modality combinations~\citep{han2024fusemoe, yun2024flex}.
Given an input mask $\mathbf{r}$, each observed modality $m\in\Omega(\mathbf{r})$ is encoded as $\mathbf{H}^m=g_m(x^m)$.
For a token representation $\mathbf{h}$, a standard sparse MoE router computes the routing distribution
\begin{equation}    
    \mathbf{q}(\mathbf{h})
    =
    \mathrm{softmax}(W_g\mathbf{h}).
\end{equation}

The MoE layer activates the top-$s$ experts according to $\mathbf{q}(\mathbf{h})$ and combines their outputs using the corresponding routing weights.
Local training typically augments the task loss with a load-balancing regularizer~\citep{yun2024flex} to prevent expert collapse and encourage balanced expert utilization.

\paragraph{Fragility of MoEs in FL under Modality Heterogeneity.}
Privacy constraints and pervasive modality incompleteness motivate multimodal FL with sparse MoE fusion for medical AI.
However, dual-level modality heterogeneity breaks the implicit assumption that sparse MoE components remain semantically comparable across clients: clients may learn different routing policies and assign different functional semantics to experts with the same index. 
To characterize the effect of modality heterogeneity on federated sparse MoEs, we analyze the routing behavior and expert representations of a standard MoE-FL baseline. 

Figures~\ref{fig:motivation_diagnostics}(a,b) focus on two representative clients with shared modality sets. Figure~\ref{fig:motivation_diagnostics}(a) reports the normalized top-$k$ activation frequency of each expert. Under the implicit cross-client comparability assumed by standard federated aggregation, the same modality type should correspond to similar expert usage across clients. However, clients exhibit distinct activation profiles, indicating that modality heterogeneity disrupts routing consistency across clients. Figure~\ref{fig:motivation_diagnostics}(b) further shows that the experts themselves become semantically misaligned. The t-SNE visualization of expert-layer outputs reveals different color and cluster structures across clients, where colors denote the top-1 routed expert. This suggests that same-index experts do not necessarily learn semantically compatible functions, and instead drift toward client-specific functional roles induced by local modality distribution. 
Such semantic misalignment diminishes the gains expected from federated aggregation.
Figure~\ref{fig:motivation_diagnostics}(c) reports the validation F1 change after aggregation relative to the pre-aggregation local model. While aggregation is consistently beneficial under balanced modality distributions, its effect becomes unreliable under heterogeneous modality distributions, sometimes yielding little improvement or even performance degradation. These results show that federated sparse MoEs are fragile under modality heterogeneity, highlighting the need for mechanisms that maintain cross-client specialization consistency during routing and aggregation. More comprehensive analyses are detailed in Appendix~\ref{app:prelim_experiments}.

\section{\system{}}\label{sec:method}
\system{} has three key components: (1)~\emph{modality-context-aware routing} guides expert selection using each token's modality identity, position, and incompleteness context, (2)~\emph{consensus-guided routing alignment} constructs reliable consensus routing anchors for shared modality patterns and aligns local batch-level routing distributions to these anchors, and (3)~\emph{client-adaptive expert aggregation} uses client-specific modality-pattern prototypes to match and aggregate functionally similar experts while preserving client-specific specialization consistency. Appendix~\ref{app:algorithm} details the algorithm.

\subsection{Modality-Context-Aware Routing}
\label{sec:method_mcr}
Standard MoE routers typically compute expert-selection scores from the token representation~\citep{fang2026aggregation, han2024fusemoe, mei2024fedmoe, yun2024flex}.
However, under dual-level modality heterogeneity, similar token representations may appear in distinct observation contexts: a modality may be observed alone, together with different complementary modalities, or under client-specific modality-availability patterns.
Since these contexts affect how a token should be interpreted, token-only routing can conflate inputs that require different expert-specialization behaviors.
We address this by introducing \emph{\textbf{M}odality-\textbf{C}ontext-aware \textbf{R}outing}~(MCR), which augments the router with each token's modality identity, observed modality mask, and token position.
Specifically, for a token representation $\mathbf{h}$ from modality $m$, observed under modality mask $\mathbf{r}$ at token position $t$, MCR first constructs a context-aware input
\begin{equation}
\mathbf{z}=\psi\left([\mathbf{h};\mathbf{e}_m;\mathbf{e}_{\mathbf{r}};\mathbf{e}_t]\right),
    \label{eq:mcr_router_input}
\end{equation}
where $\mathbf{e}_m$, $\mathbf{e}_{\mathbf{r}}$, and $\mathbf{e}_t$ are embeddings of the modality identity, modality-availability mask, and token position, respectively, and $\psi$
is a lightweight context projection network. 
The routing distribution is then
\begin{equation}
    \mathbf{q}_{\mathrm{MCR}}(\mathbf{h},m,\mathbf{r},t)
    =
    \mathrm{softmax}(W_g\mathbf{z}).
    \label{eq:mcr_router}
\end{equation}

By injecting modality context into the router input, MCR makes expert selection depend on both token content and context under which the token is observed. As a result, the router can distinguish similar tokens arising from different modality-availability patterns and produce more semantically consistent expert assignments across heterogeneous clients.

\subsection{Consensus-Guided Routing Alignment}
\label{sec:method_cra}
MCR makes routing decisions sensitive to modality context, but it does not ensure that the same modality context is routed consistently across clients.
In federated training, each client updates the router using only the modality contexts present in its local data; under inter-client modality heterogeneity, these context distributions are skewed and often incomplete.
Consequently, different clients may develop different expert preferences for the same modality pattern $p=(\mathbf{r},m)$, causing shared expert indices to lose  consistent cross-client meaning (see Figure~\ref{fig:motivation_diagnostics}).
To address this, we introduce \emph{\textbf{C}onsensus-guided \textbf{R}outing \textbf{A}lignment}~(CRA), which constructs reliable consensus anchors from cross-client routing statistics and uses them to align local routing behavior for shared modality patterns.

Formally, we define a modality pattern as $p=(\mathbf{r},m)$, denoting modality $m$ observed under mask $\mathbf{r}$. After local training, client $k$ summarizes its routing behavior for each layer $\ell$ and supported pattern $p$.
Let $\mathcal{I}_{k,p}$ denote the set of validation tokens belonging to pattern $p$.
The client computes the mean routing distribution
\begin{equation}
    \bar{\mathbf{q}}^{\ell}_{k,p}
    =
    \frac{1}{|\mathcal{I}_{k,p}|}
    \sum_{(i,t)\in\mathcal{I}_{k,p}}
    \mathbf{q}^{\ell}_{\mathrm{MCR}}(\mathbf{h}_{i,t},m,\mathbf{r}_i,t).
\end{equation}

The server combines these client-level summaries into a consensus routing anchor:
\begin{equation}
    \mathbf{a}^{\ell}_{p}
    =
    \frac{
        \sum_{k=1}^{K} w^{\ell}_{k,p}\bar{\mathbf{q}}^{\ell}_{k,p}
    }{
        \sum_{k=1}^{K} w^{\ell}_{k,p}
    },
    \label{eq:cra_anchor}
\end{equation}
where $w^{\ell}_{k,p}$ weights each client summary according to its reliability, based on the number of validation tokens with pattern $p$ and the normalized negative entropy of their token-level routing distributions.
Anchors are retained only for patterns with sufficient cross-client support, preventing rare or client-specific patterns from imposing unreliable alignment targets.

At the next communication round, the retained anchors are broadcast to clients as auxiliary routing targets.
For an anchored pattern $p$ that appears in a local training batch, let $\hat{\mathbf{q}}^{\ell}_{k,p}$ denote the batch-averaged routing distribution of the current model.
CRA minimizes
\begin{equation}
    \mathcal{L}^{k}_{\mathrm{CRA}}
    =
    \frac{1}{Z_k}
    \sum_{\ell}
    \sum_{p\in\mathcal{P}^{\ell}_k}
    \mathrm{KL}
    \left(
        \mathbf{a}^{\ell}_{p}
        \;\|\;
        \hat{\mathbf{q}}^{\ell}_{k,p}
    \right),
    \label{eq:cra_loss}
\end{equation}

where $\mathcal{P}^{\ell}_k$ is the set of anchored patterns present in the current batch at layer $\ell$, and $Z_k=\sum_{\ell}|\mathcal{P}^{\ell}_k|$.
Thus, CRA guides routing only for observable and reliably anchored patterns; when no such pattern appears in a batch, the alignment loss is zero.

CRA can provide an alignment signal only when an anchored pattern appears in a local training batch.
To increase such opportunities, we use \emph{anchor-exposure augmentation}, a training-time mask augmentation strategy that maps observed samples to feasible subpatterns of their available modalities.

For a sample with observed mask $\mathbf{r}$, define the feasible submasks
\begin{equation}    
    \mathcal{F}(\mathbf{r})
    =
    \left\{
        \mathbf{r}' :
        \Omega(\mathbf{r}') \subsetneq \Omega(\mathbf{r}),
        |\Omega(\mathbf{r}')| \geq m_{\min}
    \right\}.
\end{equation}

With probability $\rho$, we sample an augmented mask $\tilde{\mathbf{r}}\in\mathcal{F}(\mathbf{r})$ proportionally to the empirical global mask distribution $\pi_{\mathrm{G}}(\tilde{\mathbf{r}})$ and remove modalities excluded from $\mathbf{r}$; otherwise, we keep $\tilde{\mathbf{r}}=\mathbf{r}$.
If no feasible submask has positive probability under $\pi_{\mathrm{G}}$, the original mask is also retained.
By exposing clients to globally supported feasible subpatterns without introducing unobserved modalities, anchor-exposure augmentation increases the frequency with which CRA can align local routing behavior to reliable consensus anchors.

\subsection{Client-Adaptive Expert Aggregation}
\label{sec:method_cea}
MCR and CRA make expert selection context-aware and more consistent across clients, but they do not by themselves determine how expert parameters should be aggregated under modality heterogeneity.
Each client optimizes its experts on a different distribution of modality patterns, so directly averaging expert parameters can blur semantically distinct client-specific specialization behaviors.
We therefore introduce \emph{\textbf{C}lient-adaptive \textbf{E}xpert \textbf{A}ggregation}~(CEA), which personalizes expert MLPs through prototype-based functional expert matching. After local training, each client summarizes its expert-input space with pattern-level prototypes computed from validation data.
For client $k$, layer $\ell$, and pattern $p=(\mathbf{r},m)$, let $\mathbf{u}^{\ell}_{i,t}$ denote the input to the expert MLP for token $(i,t)$. Client $k$ computes
\begin{equation}    
    \bar{\mathbf{u}}^{\ell}_{k,p}
    =
    \frac{1}{|\mathcal{I}_{k,p}|}
    \sum_{(i,t)\in\mathcal{I}_{k,p}}
    \mathbf{u}^{\ell}_{i,t}.
\end{equation}

These prototypes provide compact summaries of the client's local expert-input distribution without sharing raw data. The server then uses these target-client prototypes to compare experts according to their functional behavior on shared modality patterns. For each target expert, CEA considers all source experts as candidates, allowing experts with different indices to be matched when they exhibit similar functional responses on the target client’s prototypes.
Let $g^{\ell}_{k,e}$ denote expert $e$ of client $k$ at layer $\ell$.
For target client $k$ and target expert $e$, the similarity of source expert $e'$ from client $j$ is
\begin{equation}    
    s^{\ell}_{k,e \leftarrow j,e'}
    =
    \sum_{p\in\mathcal{P}^{\ell}_{k}}
    \alpha^{\ell}_{k,p,e}
    \,
    \mathrm{cos}
    \left(
        g^{\ell}_{k,e}(\bar{\mathbf{u}}^{\ell}_{k,p}),
        g^{\ell}_{j,e'}(\bar{\mathbf{u}}^{\ell}_{k,p})
    \right),
\end{equation}

where $\mathcal{P}^{\ell}_{k}$ is the set of supported target-client patterns, and $\alpha^{\ell}_{k,p,e}$ assigns higher weight to patterns with greater prototype support and higher mean routing probability to the target expert $e$. Similarities are then converted into aggregation weights using a temperature-scaled softmax,
\begin{equation}
    \beta^{\ell}_{k,e \leftarrow j,e'}
    =
    \frac{
        \exp(s^{\ell}_{k,e \leftarrow j,e'} / \tau)
    }{
        \sum_{j'\neq k}\sum_{e''}
        \exp(s^{\ell}_{k,e \leftarrow j',e''} / \tau)
    }.
\end{equation}

Subsequently, CEA forms a personalized expert for target client $k$:
\begin{equation}    
    \tilde{g}^{\ell}_{k,e}
    =
    \eta g^{\ell}_{k,e}
    +
    (1-\eta)
    \sum_{j\neq k}\sum_{e'}
    \beta^{\ell}_{k,e \leftarrow j,e'}
    g^{\ell}_{j,e'},
\end{equation}

where $\eta$ preserves a residual contribution from the target client's original expert specialization.
The personalized experts are returned to client $k$ in the next communication round, while all non-expert components are aggregated globally.

\subsection{Federated Optimization}
\label{sec:method_training}
Appendix~\ref{app:algorithm} provides the complete round-by-round procedure.
At communication round $t$, the server sends each selected client the current globally shared components, its CEA-personalized expert parameters, and the CRA anchors constructed from the previous round, $\{\mathbf{a}^{\ell,t-1}_p\}$.
The anchors serve only as auxiliary routing targets during local optimization.

Each client then trains its local model using MCR-enabled MoE routing.
Let $\mathcal{A}_{\rho}(\cdot|\mathbf{r})$ denote the anchor-exposure augmentation distribution described above.
The local training objective is
\begin{equation}
    \mathcal{L}^{k,t}
    =
    \mathbb{E}_{(\mathbf{x},y,\mathbf{r})\sim\mathcal{D}_k}
    \mathbb{E}_{\tilde{\mathbf{r}}\sim \mathcal{A}_{\rho}(\cdot|\mathbf{r})}
    \left[
        \ell(f_{\theta}(\mathbf{x},\tilde{\mathbf{r}}),y)
    \right]
    +
    \lambda_{\mathrm{bal}}\mathcal{L}^{k,t}_{\mathrm{bal}}
    +
    \lambda_{\mathrm{CRA}}\mathcal{L}^{k,t}_{\mathrm{CRA}},
    \label{eq:medmix_local_objective}
\end{equation}

where $\mathcal{L}^{k,t}_{\mathrm{bal}}$ is the standard MoE load-balancing loss~\citep{yun2024flex} that discourages collapse onto a small subset of experts, and $\lambda_{\mathrm{bal}}$ controls its strength.
The term $\mathcal{L}^{k,t}_{\mathrm{CRA}}$ aligns current batch-level routing distributions to available consensus anchors, with weight $\lambda_{\mathrm{CRA}}$, and is zero when no anchored pattern appears in the batch. After local training, clients send model updates and pattern-level routing summaries to the server.
The server aggregates shared components, constructs next-round CRA anchors, and forms CEA-personalized experts.
Together, the components promote specialization-consistent routing and aggregation under modality heterogeneity  without sharing raw data.

\section{Experiments}\label{sec:experiments}
\paragraph{Datasets.}
We evaluate \system{} on two real-world multimodal medical datasets: ADNI~\citep{weiner2010alzheimer} and MIMIC-IV~\citep{johnson2023mimic_note,johnson2023mimic}. 
ADNI contains up to four modalities per sample, including MRI, genomics, clinical assessments, and biospecimen biomarkers, for three-class Alzheimer's diagnosis prediction. 
MIMIC-IV contains up to three EHR modalities, including laboratory measurements, discharge-note text features, and diagnosis/procedure code features, for one-year mortality prediction. 
We follow standard preprocessing from prior multimodal medical learning work~\citep{yun2024flex}. Both datasets naturally exhibit modality incompleteness; we further construct federated splits with increasing inter-client modality heterogeneity: \emph{Balanced}, \emph{Medium}, and \emph{High}. 
We further evaluate increased intra-client incompleteness and an extreme \emph{Mod-Excl.} setting, where clients have unique modality sets. 
Details are in Appendix~\ref{app:missingness_construction}.

\paragraph{Baselines.}
We compare \system{} with a broad set of baselines covering the primary method families relevant to our setting.
First, we adapt representative centralized multimodal fusion architectures to the FL setting: Attention~\citep{bahdanau2014neural}, Transformer~\citep{vaswani2017attention}, FuseMoE~\citep{han2024fusemoe}, and FlexMoE~\citep{yun2024flex}.
Second, we include FL-native incomplete-modality methods, FedDUET~\citep{orzikulova2025fedduet} and PEPSY~\citep{nguyenlearning}, which jointly address inter-client modality heterogeneity and intra-client incompleteness. Third, we include federated sparse MoE methods, FedMoE~\citep{mei2024fedmoe} and FedAlign-MoE~\citep{fang2026aggregation}, adapting their LLM fine-tuning mechanisms to our end-to-end supervised multimodal clinical prediction setting.

\paragraph{Implementation and evaluation protocol.}
All methods are implemented in PyTorch~\citep{paszke2019pytorch} and trained in the same federated simulation framework. Unless otherwise specified, each method is trained for up to 150 communication rounds with one local epoch per round and full client participation. Clients optimize the supervised task objective using Adam. We use a 70/15/15 train/validation/test split per client, select the best checkpoint according to validation F1 with smoothed early stopping, and report test F1 as the mean and standard deviation over five random seeds, using macro-F1 for ADNI and binary F1 for MIMIC-IV. For baselines, we use the best hyperparameters from the original papers when available, otherwise tuning learning rate and batch size. For \system{}, we use one sparse MoE fusion layer, gate load-balancing weight 0.01, CRA weight 0.02, anchor-exposure augmentation probability 0.3, and CEA residual weight 0.5.
All experiments were conducted using NVIDIA GeForce RTX 3090 GPUs.
Further implementation details are provided in Appendix~\ref{app:implementation_details}.

\section{Results}\label{sec:results}
\paragraph{Primary Results.}
\begin{table*}[t]
\centering
\caption{
Performance comparison under modality heterogeneity.
Results are F1 scores (macro-F1 for ADNI and binary F1 for MIMIC-IV; mean$\pm$std over five seeds).
Balanced/Medium/High denote increasing inter-client modality heterogeneity; + Intra-Client Missingness adds record-level modality dropout; Mod-Excl. denotes modality-exclusive clients.
}
\label{tab:main_results}

\begin{subtable}{\textwidth}
\centering
\resizebox{\textwidth}{!}{
\begin{tabular}{lcccccccc}
\toprule
\multirow{2}{*}{Method}
& \multicolumn{3}{c}{Inter-Client Heterogeneity}
& \multicolumn{3}{c}{+Intra-Client Missingness}
& \multicolumn{1}{c}{Extreme}
& \multirow{2}{*}{Avg.} \\
\cmidrule(lr){2-4}
\cmidrule(lr){5-7}
\cmidrule(lr){8-8}
& Balanced & Medium & High & Balanced & Medium & High & Mod-Excl. & \\
\midrule
Attention     & .507\std{.004} & .492\std{.009} & .463\std{.029} & .476\std{.006} & .471\std{.024} & .395\std{.114} & .365\std{.033} & .453 \\
Transformer   & .509\std{.009} & .498\std{.037} & .481\std{.028} & .505\std{.011} & .472\std{.018} & .501\std{.020} & .346\std{.057} & .473 \\
FuseMoE       & .520\std{.006} & \textbf{.522\std{.015}} & .487\std{.031} & .498\std{.007} & .495\std{.015} & .490\std{.026} & .382\std{.054} & .485 \\
FlexMoE       & \textbf{.568\std{.025}} & .520\std{.008} & .507\std{.025} & .515\std{.012} & .488\std{.011} & .523\std{.010} & .471\std{.045} & .513 \\
FedDUET       & .509\std{.010} & .494\std{.020} & .486\std{.005} & .464\std{.003} & .483\std{.009} & .489\std{.007} & .384\std{.023} & .473 \\
PEPSY         & .497\std{.017} & .492\std{.007} & .399\std{.023} & .476\std{.003} & .440\std{.022} & .353\std{.019} & .310\std{.097} & .424 \\
FedMoE        & .511\std{.007} & .501\std{.021} & .488\std{.044} & .454\std{.029} & .498\std{.016} & .499\std{.029} & .334\std{.043} & .469 \\
FedAlign-MoE  & .489\std{.017} & .495\std{.014} & .490\std{.027} & .479\std{.022} & .473\std{.023} & .509\std{.017} & .317\std{.039} & .464 \\
\midrule
\rowcolor{oursrow}
\textbf{\system{}}
              & .560\std{.023} & .521\std{.024} & \textbf{.546\std{.010}} & \textbf{.540\std{.012}} & \textbf{.519\std{.016}} & \textbf{.540\std{.012}} & \textbf{.481\std{.034}} & \textbf{.530} \\
\bottomrule
\end{tabular}
}
\vspace{0.05mm}
\caption*{(a) ADNI}
\label{tab:adni_main_results}
\end{subtable}

\vspace{2mm}

\begin{subtable}{\textwidth}
\centering
\resizebox{\textwidth}{!}{
\begin{tabular}{lcccccccc}
\toprule
\multirow{2}{*}{Method}
& \multicolumn{3}{c}{Inter-Client Heterogeneity}
& \multicolumn{3}{c}{+Intra-Client Missingness}
& \multicolumn{1}{c}{Extreme}
& \multirow{2}{*}{Avg.} \\
\cmidrule(lr){2-4}
\cmidrule(lr){5-7}
\cmidrule(lr){8-8}
& Balanced & Medium & High & Balanced & Medium & High & Mod-Excl. & \\
\midrule
Attention     & .596\std{.009} & .534\std{.012} & .529\std{.008} & .542\std{.007} & .538\std{.012} & .520\std{.011} & .472\std{.010} & .533 \\
Transformer   & \textbf{.616\std{.011}} & .585\std{.012} & .565\std{.012} & .564\std{.007} & .561\std{.008} & .547\std{.008} & .485\std{.015} & .561 \\
FuseMoE       & \textbf{.616\std{.006}} & .588\std{.006} & .566\std{.009} & .566\std{.008} & .563\std{.010} & .548\std{.007} & .497\std{.015} & .564 \\
FlexMoE       & .597\std{.012} & \textbf{.590\std{.008}} & .557\std{.014} & .553\std{.016} & .563\std{.014} & .543\std{.015} & .483\std{.013} & .555 \\
FedDUET       & .517\std{.006} & .507\std{.014} & .464\std{.020} & .479\std{.005} & .509\std{.022} & .448\std{.009} & .392\std{.035} & .474 \\
PEPSY         & .562\std{.003} & .578\std{.010} & .525\std{.006} & .526\std{.006} & .557\std{.018} & .499\std{.008} & .483\std{.008} & .533 \\
FedMoE        & .607\std{.018} & .574\std{.018} & .554\std{.037} & .562\std{.014} & .569\std{.009} & .538\std{.027} & .492\std{.012} & .556 \\
FedAlign-MoE  & \textbf{.616\std{.010}} & .580\std{.012} & .562\std{.011} & .564\std{.005} & .570\std{.004} & \textbf{.552\std{.008}} & .504\std{.010} & .564 \\

\midrule
\rowcolor{oursrow}
\textbf{\system{}}
              & .614\std{.005} & .587\std{.005} & \textbf{.571\std{.005}} & \textbf{.567\std{.008}} & \textbf{.576\std{.006}} & .551\std{.009} & \textbf{.505\std{.008}} & \textbf{.567} \\

\bottomrule
\end{tabular}
}
\vspace{0.05mm}
\caption*{(b) MIMIC-IV}
\label{tab:mimic_main_results}
\end{subtable}

\end{table*}

\system{} achieves the strongest average performance under modality heterogeneity on both datasets, as shown in Table~\ref{tab:main_results}. On ADNI, \system{} achieves the best average macro-F1, outperforming the strongest baseline, FlexMoE, and ranks top-2 
in every setting. The gains are most pronounced in challenging heterogeneity regimes, including High inter-client heterogeneity (0.546 vs. 0.507) and Mod-Excl. (0.481 vs. 0.471). On MIMIC-IV, \system{} also achieves the best average binary F1 (0.567), while remaining competitive across all settings and obtaining the strongest results under High inter-client heterogeneity, +Intra-Client Medium, and Mod-Excl. The relative gains are larger on ADNI because it presents a more challenging modality-heterogeneity regime, with four modalities, fewer training samples per client, and more fragmented per-silo modality-mask support. 
In contrast, MIMIC-IV has three modalities and substantially more records per client, allowing existing MoE baselines to learn comparatively stable expert specialization under moderate heterogeneity. 
This helps explain why \system{} yields especially clear improvements on ADNI, while on MIMIC-IV the gains are smaller but remain concentrated in stronger heterogeneity settings.

These results show that strong centralized multimodal MoE models are not inherently robust under federated modality heterogeneity. FuseMoE and FlexMoE remain competitive in balanced or moderate settings, but become less consistent as client modality distributions diverge, likely because routers and experts are aggregated after learning different local specialization semantics. FL-native incomplete-modality methods such as FedDUET and PEPSY explicitly address  modality incompleteness, but lack sparse expert specialization mechanisms for highly heterogeneous modality contexts. Federated MoE baselines increase expert capacity, but their aggregation mechanisms are not designed to tackle dual-level modality heterogeneity.
In contrast, \system{} explicitly promotes specialization consistency across heterogeneous clients, leading to more stable performance across datasets and modality heterogeneity regimes.

\begin{figure}[t]
    \centering
    \begin{subfigure}[t]{0.48\linewidth}
        \centering
        \includegraphics[width=\linewidth]{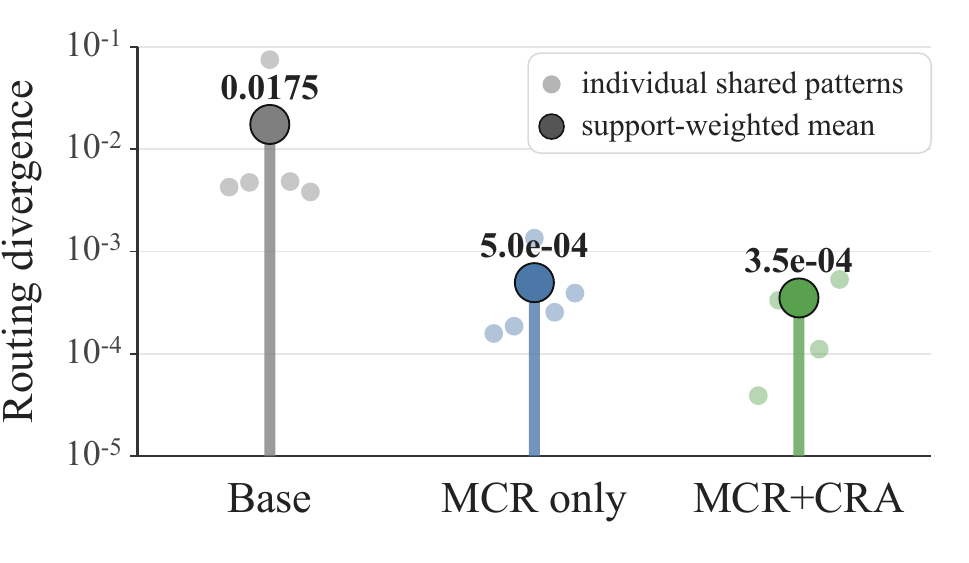}
        \caption{Cross-client routing divergence.}
        \label{fig:routing_divergence}
    \end{subfigure}
    \hfill
    \begin{subfigure}[t]{0.48\linewidth}
        \centering
        \includegraphics[width=\linewidth]{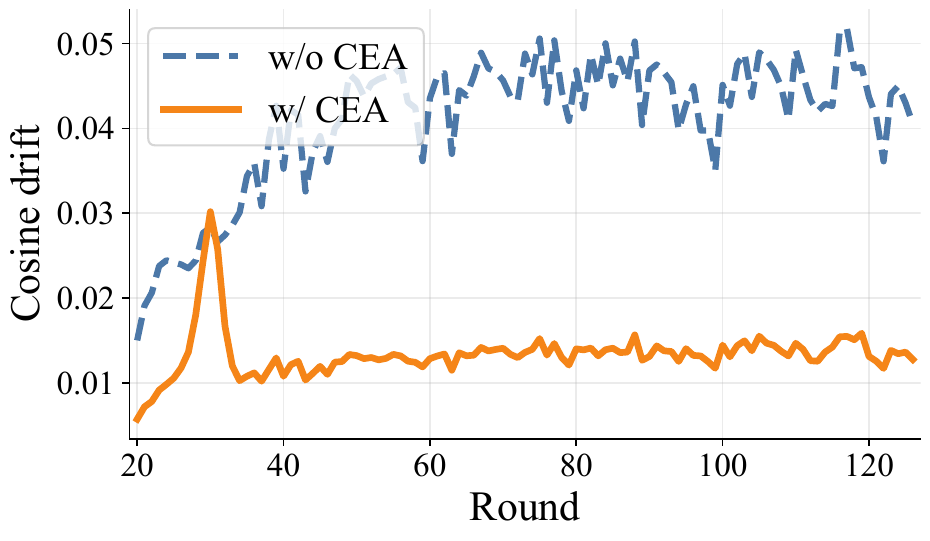}
        \caption{Prototype-weighted expert drift.}
        \label{fig:cea_drift}
    \end{subfigure}
    \caption{
    Cross-client specialization consistency analysis.
    (a) Each small dot denotes one shared modality-mask pattern, and the large dot denotes the support-weighted mean JS divergence; MCR substantially reduces routing disagreement, while CRA further improves cross-client routing consistency.
    (b) Index-wise global aggregation without CEA causes larger cosine drift from the target client's local expert, whereas CEA maintains substantially lower drift throughout training, indicating improved semantic compatibility of expert specialization.
    }
    \label{fig:specialization_consistency}
\end{figure}

\paragraph{Cross-Client Routing Alignment and Expert Compatibility.}
To examine whether \system{} improves cross-client specialization consistency, we analyze both routing behavior and expert aggregation dynamics. First, we compute the support-weighted Jensen--Shannon divergence between clients' routing distributions over shared modality-mask patterns. As shown on a log scale in Fig.~\ref{fig:specialization_consistency}(a), the baseline exhibits substantially larger cross-client routing disagreement, while MCR reduces this divergence by more than an order of magnitude. Adding CRA further lowers the divergence, suggesting that MCR improves the compatibility of routing decisions across clients and that CRA further stabilizes shared routing semantics.

Second, we evaluate whether CEA improves expert aggregation by measuring prototype-weighted functional drift between each client's local expert and the aggregated expert received after server aggregation. Fig.~\ref{fig:specialization_consistency}(b) shows that index-wise global aggregation without CEA produces steadily larger cosine drift, indicating that same-index experts are not necessarily semantically compatible across clients. In contrast, CEA keeps the personalized expert much closer to the target client's local expert throughout training, supporting its role as a function-aware aggregation mechanism that preserves specialization consistency under  modality heterogeneity.

\paragraph{System Cost.}
\label{sec:system_cost}
\begin{table*}[t]
\centering
\caption{System cost comparison. Params denote trainable per-client inference parameters. FLOPs and time are measured per sample on a balanced test split. 
Avg. F1 reports the average performance across all evaluated heterogeneity settings. 
}
\label{tab:system_cost}
\small
\setlength{\tabcolsep}{4pt}
\resizebox{\textwidth}{!}{
\begin{tabular}{lccccccccc}
\toprule
Metric & Attention & Transformer & FuseMoE & FlexMoE & FedDUET & PEPSY & FedMoE & FedAlign-MoE & \textbf{\system{}} \\
\midrule
\multicolumn{10}{l}{\textbf{ADNI}} \\
Params (M) $\downarrow$ & 3.79 & 4.17 & 10.84 & 5.02 & 4.65 & 4.46 & 10.84 & 10.84 & 4.90 \\
MFLOPs $\downarrow$    & 55.8 & 90.3 & 324.7 & 63.5 & 53.8 & 63.3 & 324.7 & 324.7 & 73.9 \\
Time (ms) $\downarrow$ & 0.82 & 0.94 & 1.85 & 2.70 & 0.72 & 1.51 & 1.92 & 1.91 & 2.37 \\
\rowcolor{oursrow} Avg. F1 $\uparrow$ & .453 & .473 & .485 & .513 & .473 & .424 & .469 & .464 & \textbf{.530} \\
\midrule
\multicolumn{10}{l}{\textbf{MIMIC-IV}} \\
Params (M) $\downarrow$ & 0.21 & 0.62 & 7.16 & 2.59 & 0.86 & 0.82 & 7.16 & 7.16 & 2.55 \\
MFLOPs $\downarrow$    & 4.9 & 25.9 & 194.6 & 9.5 & 1.9 & 9.2 & 194.6 & 194.6 & 24.6 \\
Time (ms) $\downarrow$ & 0.33 & 0.58 & 1.36 & 0.97 & 0.25 & 0.71 & 1.39 & 1.41 & 0.88 \\
\rowcolor{oursrow} Avg. F1 $\uparrow$ & .533 & .561 & .564 & .555 & .474 & .533 & .556 & .564 & \textbf{.567} \\
\bottomrule
\end{tabular}
}
\end{table*}

Table~\ref{tab:system_cost} compares \system{} with all baselines using trainable per-client inference parameters, FLOPs, wall-clock inference time, and Avg. F1 across heterogeneity settings. Cost metrics are measured per sample on a balanced test split with a fixed test-sample budget. 
\system{} achieves the highest Avg. F1 on both datasets with a moderate computational footprint. Compared with FuseMoE, FedMoE, and FedAlign-MoE, it uses about $2.2\times$ fewer parameters and $4.4\times$ fewer FLOPs on ADNI, and about $2.8\times$ fewer parameters and $7.9\times$ fewer FLOPs on MIMIC-IV. It is also comparable in size to FlexMoE while achieving higher Avg. F1 on both datasets, suggesting that its gains come from improved specialization consistency rather than increased expert capacity. Although sparse routing adds overhead beyond FLOP counts, \system{} remains in the same millisecond-scale range as the baselines and is faster than FuseMoE, FedMoE, and FedAlign-MoE on MIMIC-IV.

\paragraph{Ablation Study.}
\label{sec:results_ablation}
\begin{table}[t]
\centering
\caption{
Ablation study on ADNI.
MCR denotes modality-context-aware routing, CRA denotes consensus-guided routing alignment, and CEA denotes client-adaptive expert aggregation.}
\label{tab:ablation_adni}
\setlength{\tabcolsep}{4pt}
\renewcommand{\arraystretch}{1.08}
\resizebox{\textwidth}{!}{
\begin{tabular}{lcccccccc}
\toprule
Variant
& Balanced
& Medium
& High
& Mod-Excl.
& +Intra Bal.
& +Intra Med.
& +Intra High
& Avg. \\
\midrule
w/o CEA,CRA,MCR
& .567 & \textbf{.527} & .535 & .446 & .541 & .513 & .533 & .523 \\
w/o CEA, CRA
& \textbf{.571} & .521 & .524 & .467 & .544 & .508 & .543 & .523 \\
w/o CEA
& .554 & .524 & .544 & .467 & \textbf{.548} & .514 & \textbf{.549} & .529 \\
\midrule
\rowcolor{oursrow}
\system{} Complete
& .560 & .521 & \textbf{.546} & \textbf{.481} & .540 & \textbf{.519} & .540 & \textbf{.530} \\
\bottomrule
\end{tabular}
}
\end{table}

Table~\ref{tab:ablation_adni} ablates the main components of \system{} on ADNI.
The base variant without CEA, CRA, and MCR achieves an average F1 of 0.523, while \system{} Complete improves the average to 0.530.
Adding MCR alone (w/o CEA, CRA) improves the Mod-Excl. setting from 0.446 to 0.467, suggesting that modality-context-aware routing is especially useful when clients have strongly divergent modality patterns, though its effect is less pronounced in easier settings.
Adding CRA increases the average F1 to 0.529 and gives the best results in the +Intra Balanced and +Intra High settings.
This supports the role of consensus-guided alignment in improving cross-client routing consistency when sample-level modality incompleteness changes the modality patterns observed during local training.
Finally, adding CEA yields \system{} Complete, which achieves the best overall average and the strongest Mod-Excl. result, improving Mod-Excl. from 0.467 without CEA to 0.481. Overall, the components address complementary aspects of specialization consistency under modality heterogeneity: MCR reduces routing ambiguity, CRA improves cross-client routing consistency, and CEA preserves semantic compatibility during expert aggregation. 

\section{Discussion and Conclusion}\label{sec:conclusion}
\paragraph{Limitations and societal impacts.}
For evaluation, we construct inter-client heterogeneity and intra-client incompleteness scenarios complementary to natural modality incompleteness already present in real-world medical datasets.
This enables systematic study of modality heterogeneity, but may not capture the full complexity of real hospital deployments, where modality availability can correlate with label distributions and other site-specific clinical factors. In addition, our experiments focus on modality heterogeneity under a shared prediction task. Extending \system{} to multi-task and task heterogeneous settings remains an important direction for future work. More broadly, our results suggest that standard MoE regularization objectives such as load balancing are insufficient for federated sparse MoEs under modality heterogeneity, since balanced expert utilization alone does not guarantee semantic compatibility of expert specialization across clients. By enabling privacy-preserving, decentralized training for multimodal medical tasks under modality heterogeneity, \system{} can facilitate collaboration across institutions with different modality availability and improve model robustness for patients with partial clinical evidence. 


\paragraph{Conclusion.}
We show that federated multimodal MoEs are fragile under modality heterogeneity, as client-specific modality distributions can induce divergent routing behavior and weaken the semantic compatibility of expert specialization across clients. To address this, we propose \system{}, which promotes specialization consistency through context-aware routing, consensus-guided routing alignment, and client-adaptive expert aggregation based on functional similarity.
Experiments on ADNI and MIMIC-IV demonstrate that \system{} improves robustness across diverse modality heterogeneity and modality incompleteness settings.

\bibliographystyle{plainnat} 
\bibliography{main}    

\newpage
\appendix
\section{Algorithm}\label{app:algorithm}

\begin{algorithm}
\caption{\system{} Federated Training}
\label{alg:medmix}
\begin{algorithmic}[1]
\Require Clients $\{\mathcal{D}_k\}_{k=1}^{K}$, rounds $T$, local epochs $E$,
loss weights $\lambda_{\mathrm{bal}}, \lambda_{\mathrm{CRA}}$,
anchor-exposure rate $\rho$
\State Initialize shared params $\theta_{\mathrm{sh}}^0$ and client expert states $\{\Gamma_k^0\}_{k=1}^{K}$ from a common model
\State Initialize CRA anchors $\mathcal{B}^0=\emptyset$
\For{$t=1,\ldots,T$}
    \State Server selects clients $\mathcal{S}_t$
    \ForAll{$k\in\mathcal{S}_t$ \textbf{in parallel}}
        \State Server sends shared params $\theta_{\mathrm{sh}}^{t-1}$, CEA expert state $\Gamma_k^{t-1}$, and CRA anchors $\mathcal{B}^{t-1}$
        \State $(\theta_{\mathrm{sh},k}^{t},\Gamma_{k,\mathrm{loc}}^{t},\mathcal{R}_k^t,\mathcal{P}_k^t)
        \gets \textsc{ClientUpdate}(k,\theta_{\mathrm{sh}}^{t-1},\Gamma_k^{t-1},\mathcal{B}^{t-1})$
    \EndFor
    \State Aggregate shared params $\theta_{\mathrm{sh}}^{t}$ from $\{\theta_{\mathrm{sh},k}^{t}:k\in\mathcal{S}_t\}$ using sample-weighted FedAvg
    \State Construct CRA anchors $\mathcal{B}^{t}$ from routing summaries $\{\mathcal{R}_k^t:k\in\mathcal{S}_t\}$
    \State Form CEA-personalized expert states $\{\Gamma_k^{t}:k\in\mathcal{S}_t\}$ from local experts and prototypes $\{(\Gamma_{k,\mathrm{loc}}^{t},\mathcal{P}_k^t):k\in\mathcal{S}_t\}$
    \State Retain $\Gamma_k^{t}=\Gamma_k^{t-1}$ for clients $k\notin\mathcal{S}_t$
\EndFor
\State \Return Shared params $\theta_{\mathrm{sh}}^T$, CRA anchors $\mathcal{B}^T$, and CEA expert states $\{\Gamma_k^T\}_{k=1}^{K}$
\end{algorithmic}
\end{algorithm}

\begin{algorithm}
\caption{\textsc{ClientUpdate} for \system{}}
\label{alg:medmix_client}
\begin{algorithmic}[1]
\Require Client $k$ with $\mathcal{D}_k$, shared params $\theta_{\mathrm{sh}}$, expert state $\Gamma_k$, CRA anchors $\mathcal{B}$, local epochs $E$
\State Initialize local model by combining $\theta_{\mathrm{sh}}$ with expert params $\Gamma_k$
\For{local epoch $e=1,\ldots,E$}
    \For{minibatch $(\mathbf{x},y,\mathbf{r})$}
        \State Sample $\tilde{\mathbf{r}}\sim \mathcal{A}_{\rho}(\cdot|\mathbf{r})$ using anchor-exposure augmentation
        \State Encode modalities observed under $\tilde{\mathbf{r}}$
        \State Apply MCR routing
        \State Compute task loss $\ell(f_{\theta}(\mathbf{x},\tilde{\mathbf{r}}),y)$ and MoE load-balancing loss $\mathcal{L}_{\mathrm{bal}}$
        \State Compute CRA loss $\mathcal{L}_{\mathrm{CRA}}$ for anchored patterns present in the minibatch
        \State Update local model using
        \[
            \ell(f_{\theta}(\mathbf{x},\tilde{\mathbf{r}}),y)
            + \lambda_{\mathrm{bal}}\mathcal{L}_{\mathrm{bal}}
            + \lambda_{\mathrm{CRA}}\mathcal{L}_{\mathrm{CRA}} .
        \]
    \EndFor
\EndFor
\State Collect pattern-level routing summaries $\mathcal{R}_k$
\State Collect pattern-level expert-input prototypes $\mathcal{P}_k$
\State Split updated local model into shared params $\theta_{\mathrm{sh},k}$ and local expert params $\Gamma_{k,\mathrm{loc}}$
\State \Return $\theta_{\mathrm{sh},k}$, $\Gamma_{k,\mathrm{loc}}$, $\mathcal{R}_k$, and $\mathcal{P}_k$
\end{algorithmic}
\end{algorithm}

Routing summaries $\mathcal{R}_k$ contain support, confidence, and mean routing distributions for layer-level modality patterns $p=(\mathbf{r},m)$.
Expert prototypes $\mathcal{P}_k$ contain aggregate expert-input representations for the same patterns and are used by CEA for functional expert matching.

\section{Related Work}\label{app:related_work}

\paragraph{Multimodal Learning with Missing Modalities.}
Multimodal learning studies how models represent, align, and fuse information from multiple data sources~\cite{baltrusaitis2019multimodal}. This is especially important in medical AI, where imaging, genomics, laboratory measurements, structured records, and clinical text provide complementary evidence about a patient’s condition. In practice, however, patient records are rarely modality-complete because exams may not be ordered, measurements may be unavailable, and institutions may differ in which modalities they routinely collect. Prior work has addressed missing modalities through reconstruction-based and representation-based strategies~\cite{wu2024deep}. HeMIS embeds each available imaging modality into a shared latent space and aggregates observed modalities without synthesizing missing ones~\cite{havaei2016hemis}; Wang et al. use modality-specific teachers to learn from incomplete multimodal data while avoiding noisy imputation~\cite{wang2020multimodal}; and SMIL studies severe missingness during both training and testing with Bayesian meta-learning~\cite{ma2021smil}.

These works show that missing modalities should be handled as part of model design rather than as a preprocessing afterthought. \system{} builds on this motivation but studies a federated setting, where missingness also varies systematically across clients. Thus, incomplete modalities become not only a sample-level robustness problem, but also a source of cross-client heterogeneity that can make shared multimodal models difficult to train consistently.

\paragraph{Sparse Mixture-of-Experts for Multimodal Fusion.}
Sparse Mixture-of-Experts (MoE) models use conditional computation to increase capacity while activating only a small subset of parameters for each input. The sparsely-gated MoE layer introduced by Shazeer et al. uses a trainable gating network to select expert subnetworks on a per-example basis~\cite{shazeer2017outrageously}, while Switch Transformer simplifies sparse routing and highlights practical issues such as routing stability and expert load balancing~\cite{fedus2022switch}. This conditional structure is naturally appealing for multimodal fusion, where different modality subsets may require different fusion behavior. Recent multimodal MoE methods therefore use sparse experts to improve robustness under incomplete or variable modality observations: FuseMoE introduces a MoE Transformer for fleximodal fusion with missing modalities and irregular sampling~\cite{han2024fusemoe}, and Flex-MoE models arbitrary modality combinations using a missing-modality bank and sparse MoE routing~\cite{yun2024flex}.

These methods show that sparse MoE architectures are powerful tools for flexible multimodal fusion under missingness. However, they are primarily designed for centralized learning, where routing and expert specialization are learned under a shared data distribution. In federated multimodal learning, clients may observe different modality subsets with different frequencies, causing sparse routers and experts to specialize in client-specific ways before aggregation. \system{} addresses this federated MoE setting by coordinating routing and expert aggregation across heterogeneous clients.

\paragraph{Federated Multimodal Learning with Incomplete Modalities.}
Federated learning is a natural fit for multimodal medical AI because sensitive patient data often cannot be centralized across institutions. Multimodal federated learning (MFL), however, introduces challenges beyond standard unimodal FL: clients may differ not only in label or feature distributions, but also in which modalities are available locally~\cite{che2023multimodal}. Benchmarks such as FedMultimodal show that missing modalities are an important real-world corruption in MFL systems~\cite{feng2023fedmultimodal}. Recent methods address incomplete modalities directly in federated settings: FedMAC combines cross-modal aggregation with contrastive regularization for partial-modality missing~\cite{nguyen2024fedmac}; FedDUET studies dual-level modality heterogeneity using decoupled training and uncertainty-aware optimization~\cite{orzikulova2025fedduet}; and PEPSY learns reconfigurable representations with client-side embedding controls that encode missing-data patterns~\cite{nguyenlearning}.

These methods show that federated multimodal models must account for both inter-client modality differences and intra-client missingness. However, they primarily focus on representation alignment, reconstruction, uncertainty modeling, or personalization in non-MoE fusion architectures. In contrast, \system{} studies sparse MoE-based multimodal FL, where modality heterogeneity creates an additional failure mode: routers and experts may specialize differently across clients before aggregation. \system{} therefore complements prior incomplete-modality MFL work by explicitly coordinating routing behavior and expert aggregation under dual-level modality heterogeneity.

\paragraph{Federated Mixture-of-Experts.}
Mixture-of-experts ideas have also been explored in federated learning to address client heterogeneity. FedMix trains an ensemble of specialized models and lets each client adaptively select a user-specific subset of ensemble members, allowing clients with similar data characteristics to share statistical strength while avoiding a single global model for all clients~\cite{reisser2021federated}. This formulation uses MoE primarily as a client-level ensemble of specialized models, whereas our setting concerns sparse internal MoE layers in which multimodal tokens are routed to expert networks during fusion.

Closer to our setting, FedMoE applies sparse MoE architectures to personalized federated learning by constructing client-specific sub-MoE models from expert activation patterns and exchanging knowledge through modular aggregation and expert recommendation~\cite{mei2024fedmoe}. FedAlign-MoE further identifies routing divergence and same-index expert semantic mismatch in federated MoE training, addressing these issues through routing-distribution alignment and semantic-aware expert aggregation~\cite{fang2026aggregation}. These methods show that sparse MoE components require special care in federated optimization, but they are primarily motivated by language-model fine-tuning under general client or task heterogeneity. In contrast, \system{} studies end-to-end multimodal medical prediction under incomplete modalities, where routing and expert specialization are shaped by structured modality-availability patterns across both clients and samples.

\section{Federated Modality-Missingness Construction}
\label{app:missingness_construction}

Medical FL across institutions rarely has a perfectly aligned set of modalities.
We therefore evaluate under splits that vary two sources of modality heterogeneity: \emph{inter-client modality heterogeneity}, where clients differ in the modalities available at the institution level, and \emph{intra-client missingness}, where individual records within a client have additional missing modalities.
All split construction is monotone with respect to the dataset-provided natural modality mask: we only remove observed modalities and never add unavailable evidence.

Let $M$ denote the number of modalities.
For record $i$ at client $k$, let $\mathbf{r}^{\mathrm{nat}}_{i,k}\in\{0,1\}^{M}$ be the natural modality-availability mask from the dataset.
The final mask $\mathbf{r}_{i,k}$ always satisfies
\begin{equation}
    \mathbf{r}_{i,k} \leq \mathbf{r}^{\mathrm{nat}}_{i,k}
\end{equation}
elementwise.
Records with no remaining observed modality are removed.

\paragraph{Inter-client heterogeneity.}
We report three inter-client heterogeneity levels: \emph{Balanced}, \emph{Medium}, and \emph{High}.
The \emph{Balanced} setting uses a stratified partition that approximately preserves label and natural modality-mask distributions across clients, without adding structural modality drops.
For \emph{Medium} and \emph{High}, we simulate client-level structural modality availability using a Beta--Bernoulli process.
For each client $k$, we sample
\begin{equation}
    p_a^{(k)} \sim \mathrm{Beta}(\alpha,\beta),
\end{equation}
then draw a client-level modality-access vector
\begin{equation}
    z_{k,m} \sim \mathrm{Bernoulli}(p_a^{(k)}), \qquad m=1,\dots,M .
\end{equation}
This access vector is applied to all records in the client:
\begin{equation}
    r^{\mathrm{struct}}_{i,k,m}
    =
    r^{\mathrm{nat}}_{i,k,m} z_{k,m}.
\end{equation}
Thus, if $z_{k,m}=0$, modality $m$ is absent for every record in client $k$, mimicking institutional differences in equipment, measurement protocols, or clinical workflows.
If all modalities are dropped for a client, we randomly retain one modality to avoid a degenerate client.

\paragraph{Intra-client missingness.}
For the \emph{+Intra-Client Missingness} variants, we further apply record-level modality dropout after inter-client structural masking.
For each remaining observed modality, we independently sample
\begin{equation}
    d_{i,k,m} \sim \mathrm{Bernoulli}(p_{\mathrm{intra}})
\end{equation}
and set
\begin{equation}
    r_{i,k,m}
    =
    r^{\mathrm{struct}}_{i,k,m}(1-d_{i,k,m}).
\end{equation}
We use $p_{\mathrm{intra}}=0.3$ for the intra-client missingness variants and preserve at least one observed modality for each retained record.

\paragraph{Modality-exclusive setting.}
We also evaluate an extreme \emph{Mod-Excl.} setting, where each sufficiently frequent modality combination is assigned to a distinct client.
This creates clients with disjoint modality patterns and is used to stress-test methods under severe inter-client modality specialization.

\paragraph{Dataset-specific parameters.}
Table~\ref{tab:missingness-settings} summarizes the Beta--Bernoulli parameters used for the \emph{Medium} and \emph{High} inter-client heterogeneity levels.
The parameters are dataset-specific because ADNI has four modalities whereas MIMIC-IV has three modalities.

\begin{table}[t]
\centering
\caption{Inter-client modality-availability settings. \emph{Balanced} uses the natural stratified split without additional structural modality drops. \emph{Medium} and \emph{High} use Beta--Bernoulli client-level structural masking. The \emph{+Intra-Client Missingness} variants additionally use $p_{\mathrm{intra}}=0.3$.}
\label{tab:missingness-settings}
\setlength{\tabcolsep}{6pt}
\renewcommand{\arraystretch}{1.08}
\begin{tabular}{lclccc}
\toprule
Dataset & $M$ & Setting & Beta $(\alpha,\beta)$ & $\mathbb{E}[p_a]$ & $\mathbb{E}[\#\mathrm{avail}]$ \\
\midrule
\multirow{3}{*}{ADNI}
 & \multirow{3}{*}{4}
 & Balanced & --- & --- & $4.00$ \\
 & & Medium & $(17,3)$ & $0.85$ & $3.40$ \\
 & & High & $(6,14)$ & $0.30$ & $1.20$ \\
\midrule
\multirow{3}{*}{MIMIC-IV}
 & \multirow{3}{*}{3}
 & Balanced & --- & --- & $3.00$ \\
 & & Medium & $(15,5)$ & $0.75$ & $2.25$ \\
 & & High & $(6,14)$ & $0.30$ & $0.90$ \\
\midrule
\multicolumn{3}{l}{+Intra-Client Missingness}
& $p_{\mathrm{intra}}=0.3$ & --- & --- \\
\bottomrule
\end{tabular}
\end{table}

\section{Implementation Details}
\label{app:implementation_details}

\paragraph{Shared federated protocol.}
All federated methods are trained for up to 150 communication rounds with one local epoch per round, and all clients participate in every round.
Each client uses the Adam optimizer for local training.
We use a 70/15/15 train/validation/test split within each client.
Model selection is based on validation F1 with smoothed early stopping, using smoothing coefficient $0.1$, minimum improvement $10^{-3}$, and patience equal to $20\%$ of the training budget.
All results are reported over five seeds, $\{42,43,44,45,46\}$.

\paragraph{Missing-modality handling.}
For all methods, the dataset loader preserves the observed-modality mask for each record.
Unavailable modalities are represented by zero-valued placeholders with the correct input shape, and methods that support masking receive the observed-modality mask during fusion.
Simulated missingness splits use partition-specific labels containing the modified modality masks, while natural splits use the original dataset-level modality masks.

\paragraph{Baseline hyperparameters.}
Table~\ref{tab:baseline_hparams} summarizes the learning rates and batch sizes used for baseline methods in the federated experiments.
For standard multimodal fusion baselines, model architectures follow the corresponding centralized fusion configurations, while federated local learning rates and batch sizes are tuned separately.
FedMoE and FedAlign-MoE are adapted from their original fine-tuning setting to our end-to-end supervised multimodal prediction setting.

\begin{table}[t]
\centering
\caption{Federated baseline training hyperparameters used in our experiments.}
\label{tab:baseline_hparams}
\setlength{\tabcolsep}{5pt}
\renewcommand{\arraystretch}{1.08}
\begin{tabular}{llcc}
\toprule
Dataset & Method & Learning rate & Batch size \\
\midrule
\multirow{7}{*}{ADNI}
& Attention~\citep{bahdanau2014neural} & $4\times 10^{-5}$ & 8 \\
& Transformer~\citep{vaswani2017attention} & $1\times 10^{-4}$ & 8 \\
& FuseMoE~\citep{han2024fusemoe} & $1\times 10^{-4}$ & 16 \\
& FlexMoE~\citep{yun2024flex} & $1\times 10^{-4}$ & 8 \\
& FedDUET~\citep{orzikulova2025fedduet} & $1\times 10^{-4}$ & 16 \\
& PEPSY~\cite{nguyenlearning} & $4\times 10^{-5}$ & 8 \\
& FedMoE~\citep{mei2024fedmoe} & $1\times 10^{-4}$ & 16 \\
& FedAlign-MoE~\citep{fang2026aggregation} & $1\times 10^{-4}$ & 16 \\

\midrule
\multirow{7}{*}{MIMIC-IV}
& Attention~\citep{bahdanau2014neural} & $1\times 10^{-4}$ & 16 \\
& Transformer~\citep{vaswani2017attention} & $4\times 10^{-5}$ & 8 \\
& FuseMoE~\citep{han2024fusemoe} & $1\times 10^{-4}$ & 16 \\
& FlexMoE~\citep{yun2024flex} & $4\times 10^{-5}$ & 16 \\
& FedDUET~\citep{orzikulova2025fedduet} & $4\times 10^{-5}$ & 16 \\
& PEPSY~\citep{nguyenlearning} & $1\times 10^{-4}$ & 16 \\
& FedMoE~\citep{mei2024fedmoe} & $1\times 10^{-4}$ & 16 \\
& FedAlign-MoE~\citep{fang2026aggregation} & $1\times 10^{-4}$ & 16 \\
\bottomrule
\end{tabular}
\end{table}

\begin{table}[t]
\centering
\caption{\system{} model and training hyperparameters.}
\label{tab:medmix_hparams}
\setlength{\tabcolsep}{5pt}
\renewcommand{\arraystretch}{1.08}
\begin{tabular}{lccccccc}
\toprule
Dataset & Experts & Top-$k$ & Heads & Layers & Classifier hidden/layers & LR & Batch size \\
\midrule
ADNI & 16 & 2 & 4 & 1 & 256 / 2 & $1\times 10^{-4}$ & 8 \\
MIMIC-IV & 32 & 4 & 4 & 1 & 320 / 3 & $4\times 10^{-5}$ & 16 \\
\bottomrule
\end{tabular}
\end{table}

\paragraph{\system{} hyperparameters.}
Table~\ref{tab:medmix_hparams} summarizes the main \system{} hyperparameters.
For both datasets, we use one sparse MoE fusion layer, a single router, gate load-balancing weight $0.01$, MCR modality/mask/position embedding dimensions of $32/64/16$, and router hidden dimension $256$.
CRA uses minimum per-client pattern support $160$, quorum size $2$, alignment weight $0.02$, and starts from round 2.
Anchor-exposure augmentation uses targeted modality dropout with probability $0.3$ and keeps at least one modality.
CEA uses cross-expert matching, residual self-weight $0.5$, temperature $0.2$, unsupported-source weight $0.5$.

\section{Additional Preliminary Experiment Details}\label{app:prelim_experiments}

We provide additional details and results for the preliminary diagnostics in Section~\ref{sec:background}. All experiments are conducted on ADNI, and use the same sparse MoE fusion backbone as \system{} with standard FedAvg aggregation. Unless otherwise stated, statistics are computed from final local models before server aggregation and on each client's validation set.

\begin{figure*}[t]
  \centering
  \includegraphics[width=0.92\textwidth]{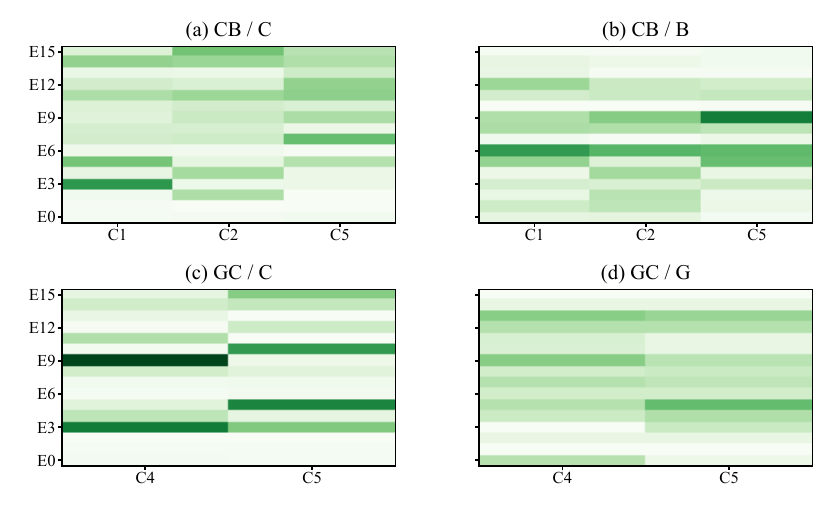}
  \caption{\textbf{Additional routing-divergence examples.}
  Normalized top-2 expert activation frequencies for four different observed modality patterns. Here, C, B, and G denote clinical, biospecimen, and genomic modalities; for example, CB / C refers to the clinical modality token when only clinical and biospecimen modalities are observed for the sample. Darker green indicates more frequent expert selection. When routing the same modality pattern, clients activate different expert subsets, extending Figure~\ref{fig:motivation_diagnostics}(a).}
  \label{fig:appendix_router_examples}
\end{figure*}

\paragraph{Router top-$k$ activation frequencies.}
For each modality pattern $p=(\mathbf{r},m)$, we count how often each expert is selected by the top-2 router and normalize counts within each client. Figure~\ref{fig:appendix_router_examples} shows additional examples where clients show diverging routing distributions even for the same modality patterns.

\begin{figure*}[t]
  \centering
  \includegraphics[width=\textwidth]{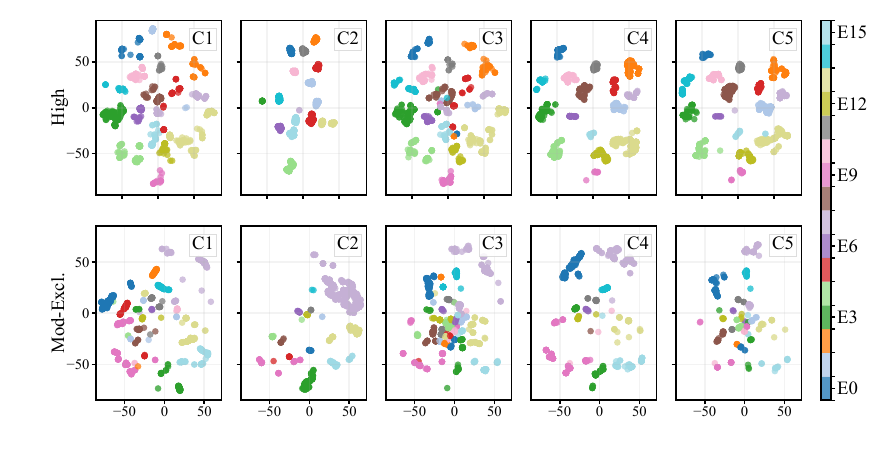}
  \caption{\textbf{Expert misalignment across clients.}
  t-SNE visualizations of final-round expert outputs for all clients in the High and Mod-Excl. heterogeneity settings. Each row uses a shared embedding across clients, and colors denote top-1 routed experts. Structure varies across clients, extending Figure~\ref{fig:motivation_diagnostics}(b).}
  \label{fig:appendix_expert_semantics}
\end{figure*}

\paragraph{Expert-output t-SNE visualizations.}
We collect final-round expert-output vectors and color each point by its top-1 routed expert. For each row in Figure~\ref{fig:appendix_expert_semantics}, one t-SNE embedding is fit jointly across all five clients in that split and then faceted by client.

\begin{figure*}[t]
  \centering
  \includegraphics[width=0.78\textwidth]{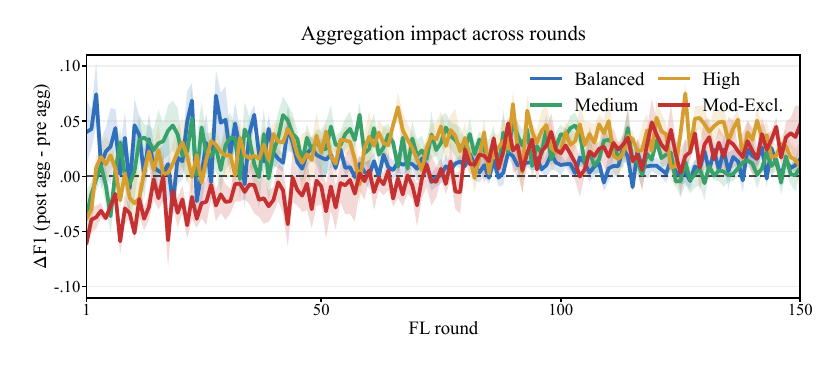}
  \caption{\textbf{Aggregation impact over training.}
  Round-wise aggregation impact, $\Delta\mathrm{F1}=\mathrm{F1}_{\mathrm{post}}-\mathrm{F1}_{\mathrm{pre}}$, for four different heterogeneity settings.
  Lines show means over clients and seeds. Shaded regions denote standard error over seeds.
  Negative values indicate that aggregation hurts the just-trained local model.}
  \label{fig:appendix_aggregation_dynamics}
\end{figure*}

\paragraph{Aggregation impact over rounds.}
At each round $t$, we evaluate client $k$ before and after FedAvg aggregation on the same client validation set:
\begin{equation}
  \Delta\mathrm{F1}_{k,t}
  =
  \mathrm{F1}^{\mathrm{post}}_{k,t}
  -
  \mathrm{F1}^{\mathrm{pre}}_{k,t}.
\end{equation}
Figure~\ref{fig:appendix_aggregation_dynamics} plots the round-wise mean over clients and five seeds on four different heterogeneity settings.

\clearpage
\section{License of Assets}\label{app:license_assests}
This work uses two restricted-access third-party clinical research datasets:
the Alzheimer's Disease Neuroimaging Initiative (ADNI)~\citep{weiner2010alzheimer} dataset and MIMIC-IV~\citep{johnson2023mimic_note, johnson2023mimic}.
We did not create, redistribute, sublicense, or release participant-level data
from either resource. All experiments were conducted by authorized users under
the applicable data access agreements. ADNI data were obtained from the ADNI database through the LONI Image and Data Archive (IDA). Only aggregate results are reported in this paper, and no ADNI participant-level data are included in the paper, supplementary material, code release, or model artifacts. MIMIC-IV was accessed through PhysioNet. We report only aggregate, de-identified experimental results. No MIMIC-IV records, tables, notes, or participant-level derivatives are redistributed with this submission. Users who wish to
reproduce experiments requiring these datasets must independently obtain access
from ADNI/LONI and PhysioNet and comply with the respective data use
agreements.


\end{document}